\documentclass[letterpaper]{article} 
\usepackage{aaai2026}  
\usepackage{times}  
\usepackage{helvet}  
\usepackage{courier}  
\usepackage[hyphens]{url}  
\usepackage{graphicx} 
\usepackage{natbib}  
\usepackage{caption} 
\usepackage{algorithm}
\usepackage{algorithmic}
\usepackage{multirow}
\usepackage{multicol}

\usepackage{amssymb}
\usepackage{amsmath}
\usepackage{subcaption}
\usepackage{cleveref}
\usepackage{mathtools}
\usepackage{array}
\usepackage{booktabs}
\usepackage{xcolor}
\usepackage{newfloat}
\usepackage{listings}
\DeclareCaptionStyle{ruled}{labelfont=normalfont,labelsep=colon,strut=off} 
\floatstyle{ruled}
\newfloat{listing}{tb}{lst}{}
\floatname{listing}{Listing}
\title{Adaptive Theory of Mind for LLM-based Multi-Agent Coordination}
\author{
    Chunjiang Mu\textsuperscript{\rm 1,\rm2,}\thanks{This work was done during his internship at Shanghai Artificial Intelligence Laboratory.}, 
    Ya Zeng\textsuperscript{\rm 1},
    Qiaosheng Zhang\textsuperscript{\rm 2}, 
    Kun Shao\textsuperscript{\rm 3}, 
    Chen Chu\textsuperscript{\rm 4}, 
    Hao Guo\textsuperscript{\rm 5}, 
    Danyang Jia\textsuperscript{\rm 1}, 
    Zhen Wang\textsuperscript{\rm 1,}\thanks{Corresponding author.}, 
    Shuyue Hu\textsuperscript{\rm 2,\textdagger}
}
\affiliations{
    \textsuperscript{\rm 1}School of Cybersecurity, Northwestern Polytechnical University\\
    \textsuperscript{\rm 2}Shanghai Artificial Intelligence Laboratory\\
    \textsuperscript{\rm 3}Huawei Noah's Ark Lab\\
    \textsuperscript{\rm 4}School of Statistics and Mathematics, Yunnan University of Finance and Economics\\
    \textsuperscript{\rm 5}QiYuan Lab\\
    w-zhen@nwpu.edu.cn, 
    hushuyue@pjlab.org.cn

}

\usepackage{bibentry}

\begin{document}

\maketitle

\begin{abstract}
Theory of Mind (ToM) refers to the ability to reason about others’ mental states, and higher-order ToM involves considering that others also possess their own ToM.
Equipping large language model (LLM)-driven agents with ToM has long been considered to improve their coordination in multi-agent collaborative tasks.
However, we find that misaligned ToM orders---mismatches in the depth of ToM reasoning between agents---can lead to insufficient or excessive reasoning about others, thereby impairing their coordination.
To address this issue, we design an adaptive ToM (A-ToM) agent, which can align in ToM orders with its partner. 
Based on prior interactions, the agent estimates the partner’s likely ToM order and leverages this estimation to predict the partner’s action, thereby facilitating behavioral coordination.
We conduct empirical evaluations on four multi-agent coordination tasks: a repeated matrix game, two grid navigation tasks and an Overcooked task. 
The results validate our findings on ToM alignment and demonstrate the effectiveness of our A-ToM agent. 
Furthermore, we discuss the generalizability of our A-ToM to non-LLM-based agents, as well as what would diminish the importance of ToM alignment.
\end{abstract}

\begin{links}
\link{Code}{https://github.com/ChunjiangMonkey/Adaptive-ToM}
\end{links}

\section{1. Introduction}
Multi-agent coordination involves the precise alignment of actions among multiple agents to enable effective joint behavior, and is widely applied in areas such as autonomous driving \citep{zhang2024multi}, swarm robotics \citep{kegeleirs2025towards}, and distributed control \citep{ge2025distributed}.
A key challenge in this area is zero-shot coordination, where agents need to coordinate with previously unseen partners without prior joint training or communication \citep{hu2020other}.
Large language models (LLMs) have been widely used to construct zero-shot coordination agents, as they possess strong decision-making and generalization capabilities, and can be deployed without task-specific training~\citep{agashe2023llm, zhang2024proagent, liu2024llm}.

Effective coordination with unseen partners requires the ability to model and anticipate their behavior. Recent research has incorporated explicit Theory of Mind (ToM) into the architecture of LLM-based agents, enabling them to model others by reasoning about others’ beliefs, desires, and intentions \citep{li2023theory, agashe2023llm}. 
ToM-based workflow designs and prompting techniques have demonstrated clear effectiveness and become an important component in the design of agents in multi-agent problems. More generally, since other agents may also possess ToM capabilities, it is necessary to equip LLM-based agents with higher-order ToM to reason about others’ reasoning (e.g., ``I believe that you believe...”) \citep{de2014agent, wellman2018theory}.
However, it has been found that higher ToM orders do not necessarily improve performance, in either cooperative or competitive multi-agent tasks \citep{li2023theory, shao2024cognitive, zhang2025k}.
Prior work empirically attributes the performance drop either to the limited ability of LLMs to perform higher-order ToM reasoning, or to over-reasoning introduced by higher-order ToM itself.

In this paper, we highlight a deeper underlying cause for the performance drop observed between agents with ToM---\textbf{the misalignment of their ToM orders}.
According to the definition of the ToM order, an agent with $k$th-order ToM aligns best with agents of $(k{-}1)$-th or $(k{+}1)$-th order; otherwise, the mismatch may lead to either insufficient or excessive reasoning.
For instance, consider two cars driving toward each other on a narrow road. If both drivers attempt to avoid a collision by swerving to the same side, an accident can still occur. This is a typical case where the misalignment of ToM (both 1st-order ToMs in this case) leads to serious consequences. We find that such misalignment has a significant impact on coordination between LLM-based agents by experiments. 

To address this issue, we propose the first adaptive ToM agent (A-ToM) driven by LLMs, which estimates its partner's ToM order in real time and selects actions that structurally align with it. 
The A-ToM agent consists of multiple hypothetical agents, each representing a order of ToM. During real-time interactions, the A-ToM agent selects one of candidate actions---generated by these hypothetical agents---as its prediction of the partner’s action. This selection is guided by the historical prediction accuracies of the hypothetical agents.

We model this process as an \textbf{Expert Advice} problem \citep{cesa1997use}, and we solve it using online learning algorithms that have theoretical performance guarantees. Finally, our A-ToM agents selects actions from the available action set that can coordinate with the prediction action to interact with the actual partner.
Through experiments on a repeated matrix game, two grid world navigation tasks and an Overcooked task, we validate our A-ToM agent can robustly coordinate with different types of partners. Overall, our contributions are as follows:
\begin{itemize}
\item We identify alignment in ToM orders between agents as a critical factor for achieving successful coordination.
\item We develop an A-ToM agent for zero-shot coordination which can align with the partner's ToM order in real time.
\item We validated the correctness of our findings and the effectiveness of our A-ToM agent across multiple coordination tasks. Furthermore, we analyze the generalization of our findings and A-ToM agent. 
\end{itemize}

\section{2. Related Work}

\subsubsection{Large Language Model-Based Agent.}
Equipped with various modules such as perception, memory, and controller, LLM-based agents have proven successful in addressing difficult tasks across multiple domains, such as robotic control \citep{brohan2023can, wu2023tidybot}, industrial automation \citep{xia2023towards}, GUI operation \citep{gurreal, yan2023gpt}, and playing open-world games \citep{wang2023voyager, wang2023describe}. 
Moreover, multiple LLM-based agents can collaborate to accomplish large-scale task like software development \citep{hong2023metagpt, qian2024chatdev} and social simulation \citep{park2023generative}. 
The above work demonstrates that LLM-based agents can make reliable decisions in both single-agent and multi-agent tasks. 
Therefore, we leverage the LLM-based agent as a rational decision-maker to investigate the impact of ToM alignment between rational collaborators. 
In addition, due to their promising generalization ability, using of LLM-based agents eliminates the need to design decision rules or train agents from scratch for each task.

\subsubsection{Enhancing Multi-Agent Collaboration Through Theory of Mind.}
ToM is the capacity of human to reason about the beliefs, desires, and intentions of others, which is crucial in human social interactions. 
In multi-agent collaboration, explicitly equipping AI agents—including LLM-based agents—with ToM enables them to infer others’ hidden states and predict their behavior, thereby enhancing communication efficiency between agents \citep{wangtom2c, zhu2021few, li2023theory}, overcoming challenges of partial observability in the environment \citep{fuchs2021theory, cross2024hypothetical}, and improving coordination among agents \citep{wu2021too, agashe2023llm, zhang2024proagent}.
However, some studies have shown that equipping agents with higher-order ToM does not always lead to the expected improvements \citep{li2023theory, shao2024cognitive}.
In this paper, we investigate the critical role of ToM reasoning depth alignment in facilitating collaboration between LLM-based agents. 
Our work is primarily inspired by prior research on inferring others' ToM \citep{yoshida2008game, de2013much} and dynamically matching partners for agents with specific ToM orders \citep{shao2024cognitive}.

\section{3. Method}

\subsection{3.1 Problem Formulation}
We consider a fully cooperative decision-making problem involving two agents within a given Markovian environment, where the actions of agents require coordination to achieve an optimal outcome. 
The environment can be formalized as a tuple $\mathcal{M} = \langle \mathcal{S}, \mathcal{A}_1, \mathcal{A}_2, \mathcal{T}, R, \gamma \rangle$.
$\mathcal{S}$ is the shared state space. $\mathcal{A}_1$ and $\mathcal{A}_2$ are the action spaces of the two agents. $\mathcal{T}: \mathcal{S} \times \mathcal{A}_1 \times \mathcal{A}_2 \rightarrow \Delta(\mathcal{S})$ is the transition function.
$R: \mathcal{S} \times \mathcal{A}_1 \times \mathcal{A}_2 \rightarrow \mathbb{R}$ is a shared reward function that ensures that both agents receive the same reward. 
$\gamma \in [0,1)$ is the discount factor.
Denote the policies of two agents by $\pi_1: \mathcal{S} \rightarrow \Delta(\mathcal{A}_1)$ and $\pi_2: \mathcal{S} \rightarrow \Delta(\mathcal{A}_2)$. 
The joint policy of two agents is denoted as $\mathbb{\pi} = (\pi_1, \pi_2)$. 

Although LLM-based agents do not require a reward function for training, we still use the discounted expected return (i.e., the value function) to define the \textit{rationality} of LLM-based agents. 
At each step, two rational LLM-based agents aim to select the joint actions $\mathbf{a}^* \coloneqq (a_1^*, a_2^*)$ that maximize their joint value function:
\begin{equation}
    \mathbf{a}^* = \arg\max_{\mathbf{a} \in \mathcal{A}_1 \times \mathcal{A}_2} Q_{\boldsymbol{\pi}}(s, \mathbf{a}),
\end{equation}
where
\begin{equation}
Q_{\boldsymbol{\pi}}(s, \mathbf{a}) = \mathbb{E}_{\boldsymbol{\pi}} \left[ \sum_{t=0}^{\infty} \gamma^t \, R(s_t, \mathbf{a}_t) \,\middle|\, s_0 = s,\ \mathbf{a}_0 = \mathbf{a} \right].
\end{equation}
In some states, there are multiple optimal joint actions satisfying $\mathbf{a}^{*,1}=\mathbf{a}^{*,2}=\cdots=\max Q_{\boldsymbol{\pi}}(s, \mathbf{a})$. Two agents must coordinate to agree on the same optimal joint action. However, it can be particularly challenging in the absence of communication or prior agreement \citep{boutilier1999sequential}.


\subsection{3.2. ToM Modeling}

The order of ToM is the depth of recursive reasoning an agent uses to model its partner’s behavior. For convenience, we refer to an agent with $k$-th order ToM as a ToM-$k$ agent. We now define the decision-making process of an agent $i$ with different orders of ToM reasoning as follows. without loss of generality, we assume $i = 2$.
\subsubsection{ToM-0 agent.}  A ToM-0 agent treats its partner as part of the environment state. Its decision depends solely on the environment state:
\begin{equation}
    \pi_i^{(0)}(s)\coloneqq \arg\max_{a\in \mathcal{A}_i} Q_{\boldsymbol{\pi}}(s, a).\label{pi0}
\end{equation}
\subsubsection{ToM-1 agent.} A ToM-1 agent assumes that its partner $j$ is a ToM-0 agent. Its first-order belief $b_i^{(1)}$ is partner $j$'s predicted action $a_j^{\text{pred}}$:
\begin{equation}
    b_i^{(1)}\coloneqq a_j^{\text{pred}}, \quad \text{where } a_j^{\text{pred}}=\pi_j^{(0)}(s).\label{b1}
\end{equation}
The agent then selects the action that best coordinates with $a_j^{\text{pred}}$:
\begin{equation}
    \pi_i^{(1)}(s, b_i^{(1)})\coloneqq\arg\max_{a\in \mathcal{A}_i} Q_{\boldsymbol{\pi}}(s,a_j^{\text{pred}},a).\label{pi1}
\end{equation}
\subsubsection{ToM-2 agent.} Similar to a ToM-1 agent, a ToM-2 agent first infers the partner's action $a_j^{\text{pred}}$ as its second-order belief. Differently, the ToM-2 agent thinks that its partner $j$ is a ToM-1 agent and agent $j$ thinks agent $i$ is a ToM-0 agent:
\begin{equation}
    b_i^{(2)}\coloneqq a_j^{\text{pred}}, \quad \text{where } a_j^{\text{pred}}=\pi_j^{(1)}(s, b_j^{(1)}).
\end{equation}
Here, $b_j^{(1)}$ can be computed in the same manner as in \cref{pi0}, \cref{b1}  and \cref{pi1}. The subscript $j$ in $b_j^{(1)}$ indicates that this belief corresponds to agent $i$'s prediction---made from agent $j$'s perspective---about the action agent $i$ will take.
Then, the ToM-2 agent's policy is to take an action that best responds to $b_i^{(2)}$:
\begin{equation}
    \pi_i^{(2)}(s, b_i^{(2)})\coloneqq\arg\max_{a\in \mathcal{A}_i} Q_{\boldsymbol{\pi}}(s,a_j^{\text{pred}},a).\label{pi2}
\end{equation}

\subsubsection{ToM-$k$ agent.} More generally, the policy of a ToM-$k$ agent $(k>0)$ is defined recursively as:
\begin{equation}
    b_i^{(k)}\coloneqq a_j^{\text{pred}}, \quad \text{where } a_j^{\text{pred}}=\pi_j^{(k-1)}(s, b_j^{(k-1)}).
\end{equation}
\begin{equation}
    \pi_i^{(k)}(s, b_i^{(k)})\coloneqq\arg\max_{a\in \mathcal{A}_i} Q_{\boldsymbol{\pi}}(s,a_j^{\text{pred}},a).\label{pik}
\end{equation}
In this work, we focus to $k \leq 2$, since higher-order ToM imposes significant cognitive burdens and empirical studies suggest that humans typically reason about others only up to the second-order ToM \citep{camerer2004cognitive,devaine2014social}.
The recursive structure of ToM reasoning implies that a ToM-$k$ agent assumes its partner to be a ToM-$(k{-}1)$ agent. \textit{Naturally, such agents are more likely to coordinate successfully when interacting with ToM-$(k{-}1)$ or ToM-$(k{+}1)$ partners.} We refer to this compatibility as \textbf{aligned ToM orders}.
Based on this insight, we argue that agents should focus on understanding how their partner thinks, rather than just reacting to what their partner does. Adapting to the partner’s ToM order is a more effective way to achieve coordination.
\begin{figure}[t]
    \centering
    \includegraphics[width=0.4\textwidth]{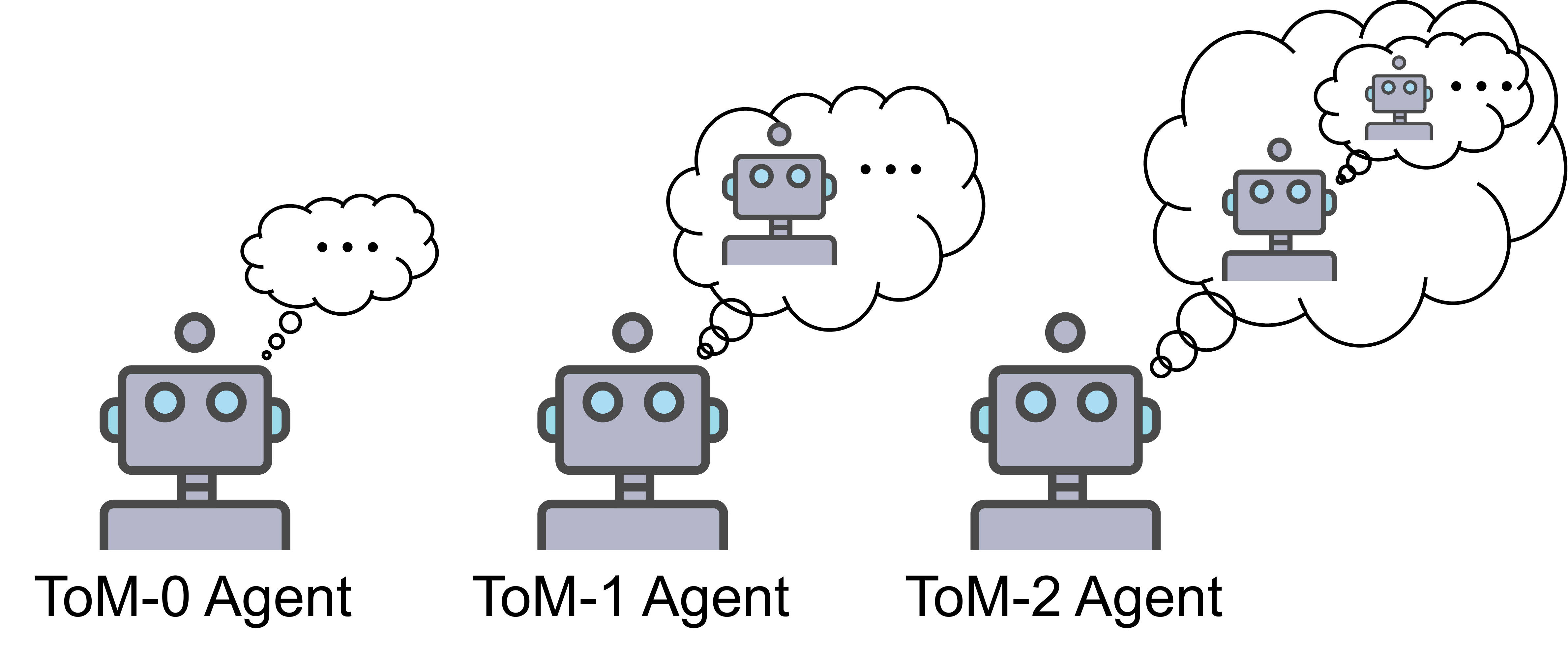}
    \caption{Illustration of ToM reasoning of different orders.}
    \label{fig:tom_level}
\end{figure}

\subsection{3.3. ToM Alignning}
We now introduce our adaptive ToM agent (A-ToM agent) that can dynamically estimate the ToM order of the partner during the interaction. Specifically, we formulate the ToM order alignment problem as an online \textbf{expert advice} problem, where each ToM-$k$ policy $\pi_j^{(k)}$ is treated as an expert \citep{cesa1997use}.

The A-ToM agent maintains a set of hypothetical agents with distinct ToM order $[\pi_j^{(k)}]_{k \in \{0,1,2\}}$ along with their corresponding cumulative losses (or weights). 
The agent continuously updates these weights based on interaction outcomes to better estimate the true ToM order of partner $j$.
The update process is as follows: 1) using hypothetical agent to generate a set of candidate actions $[\hat{a}_j^{(k)}]_{k \in \{0,1,2\}}$; 2) selecting one candidate action as the predicted partner action $\hat{a}_j$, based on corresponding hypothetical agents' historical prediction accuracies; 3) selecting a response action coordinating with the predicted partner action; and (4) observing the actual partner action and updating the prediction accuracy of each hypothetical agent.

The online learning mechanisms for Steps 2) and 4) are determined by the specific expert advice algorithm. In this work, we consider two such algorithms: Follow-the-Leader (FTL) \citep{kalai2005efficient} and Hedge \citep{freund1997decision}, which are shown in detail in \cref{alg:ftl} and \cref{alg:hedge}, respectively. FTL achieves a regret bound of $\mathcal{O}(\log T)$ in stable settings, making it suitable for coordination with partners of fixed ToM orders. 
Hedge, in contrast, maintains a soft expert weight distribution over ToM orders, enabling it to handle uncertainty and adapt to non-stationary behavior with a worst-case regret bound of $\mathcal{O}(\sqrt{T\log N})$. 
Here, $T$ denotes the total number of executions of the algorithm and $N$ is the number of experts. In this work, $N = 3$ for we consider three ToM orders. 

\textit{Conceptually, our ToM alignment framework transforms the original coordination problem in the policy space into a alignment problem in ToM-order space, thereby reducing the dimensionality and structural complexity of coordination. } From a perspective of learning, our method leverages the reasoning capabilities of LLMs by assigning them learning tasks at an abstraction order that aligns naturally with their strengths, rather than burdening them with low-level, fine-grained details.

\begin{algorithm}[h]
\caption{ToM Alignment via Follow-the-Leader (FTL)}
\label{alg:ftl}
\begin{algorithmic}[1]
\STATE \textbf{Input:} The environment $\mathcal{M}$, ego agent $\pi_i$, candidate ToM orders $\mathcal{K} = \{0,1,2\}$, hypothetical ToM agents $\{\pi_j^{(k)}\}_{k \in \mathcal{K}}$
\STATE Initialize cumulative loss $L^{(k)} \gets 0$ for all $k \in \mathcal{K}$
\STATE Initialize $t \gets 1$
\WHILE{$\mathcal{M}$ is not terminated }
    \STATE Observe current state $s$
    \FOR{each $k \in \mathcal{K}$}
        \STATE $\hat{a}_j^{(k)} \gets \pi_j^{(k)}(s, b_j^{(k)})$
    \ENDFOR
    \STATE $\hat{k} \gets \arg\min_{k \in \mathcal{K}} L^{(k)}$, $a_j^{\text{pred}} \gets \hat{a}_j^{(\hat{k})}$
    \STATE Acting: $a_i^t \gets \pi_i(s,a_j^{\text{pred}})$
    \STATE $t \gets t+1$
    \STATE Observing true partner's action $a_j^{t-1}$
    \FOR{each $k \in \mathcal{K}$}
        \IF{$\hat{a}_j^{(k)} \ne a_j^{t-1}$}
             \STATE Update loss: $L^{(k)} \gets L^{(k)} + 1$
        \ENDIF
    \ENDFOR
\ENDWHILE
\end{algorithmic}
\end{algorithm}

\begin{algorithm}[h]
\caption{ToM Alignment via Hedge}
\label{alg:hedge}
\begin{algorithmic}[1]
\STATE \textbf{Input:} The environment $\mathcal{M}$, ego agent $\pi_i$, candidate ToM orders $\mathcal{K} = \{0,1,2\}$, hypothetical ToM agents $\{\pi_j^{(k)}\}_{k \in \mathcal{K}}$, learning rate $\eta = 1$
\STATE Initialize expert weight $w^{(k)} \gets 1$ for all $k \in \mathcal{K}$
\STATE Initialize $t \gets 1$
\WHILE{$\mathcal{M}$ is not terminated }
    \STATE Observe current state $s$
    \FOR{each $k \in \mathcal{K}$}
        \STATE Normalize expert weight: $P^{(k)} \gets \frac{w^{(k)}}{\sum_{k'\in \mathcal{K}} w^{(k')}}$
        \STATE $\hat{a}_j^{(k)} \gets \pi_j^{(k)}(s, b_j^{(k)})$
    \ENDFOR
    \STATE $\hat{k} \sim  P^{(k)}$, $a_j^{\text{pred}} \gets \hat{a}_j^{\hat{k}}$
    \STATE Acting: $a_i^t \gets \pi_i(s,a_j^{\text{pred}})$
    \STATE $t \gets t+1$
    \STATE Observing true partner's action $a_j^{t-1}$
    \FOR{each $k \in \mathcal{K}$}
        \IF{$\hat{a}_j^{(k)} \ne a_j^t$}
            \STATE $\ell^{(k)} \gets 1$
        \ELSE
            \STATE $\ell^{(k)} \gets 0$
        \ENDIF
        \STATE Update expert weight: $w^{(k)} \gets w^{(k)} \cdot \exp(-\eta \cdot \ell^{(k)})$
    \ENDFOR
\ENDWHILE
\end{algorithmic}
\end{algorithm}



\subsection{3.4. LLM-Based Agent Implementation}

We use LLMs to construct agents with fixed ToM orders and the A-ToM agent. Each LLM-based agent consists of four modules: a state encoding module, a ToM module, a decision module, and an action controller. The state encoding module converts the structured environment state into a natural language description. The ToM module predicts the partner’s action, where we recursively construct hypothetical agents with different ToM orders. For a ToM-$k$ agent, the ToM module infers the partner's behavior by invoking a ToM-$(k{-}1)$ hypothetical agent. For our A-ToM agent, the ToM module includes three hypothetical agents: a ToM-0 agent, a ToM-1 agent, and a ToM-2 agent. The decision module takes as input both the state description and the predicted partner action from the ToM module, and outputs the agent's action. The action controller converts natural language actions produced by the LLM into executable actions in the environments.
In summary, we follow the two-stage design of LLM-based agents with ToM: first, the LLM predicts the partner's behavior; then, this predicted behavior is incorporated into the LLM's input to inform action selection \citep{agashe2023llm, zhang2024proagent}.
 Given that state-of-the-art LLMs already provide sufficiently reliable output quality for most planning tasks, we do not introduce any additional output verification component in practice, which also helps reduce inference latency and system complexity.


\begin{table*}[h]
\centering
\small
\begin{tabular}{llccccc}
\toprule
\multirow{2}{*}{\textbf{Alignment}}&\multirow{2}{*}{\textbf{Agent Profile}}& \multicolumn{1}{c}{\textbf{Memory-1}} & \multicolumn{1}{c}{\textbf{Memory-N}} & \multicolumn{1}{c}{\textbf{Game 1}} & \multicolumn{1}{c}{\textbf{Game 2}}&\multicolumn{1}{c}{\textbf{Overcooked}}  \\
\cmidrule(r){3-3}
\cmidrule(r){4-4}
\cmidrule(r){5-5}
\cmidrule(r){6-6}
\cmidrule(r){7-7}
& & Point $\uparrow$ & Point $\uparrow$ & Time $\downarrow$& Time $\downarrow$ & Time $\downarrow$\\
\midrule
\multirow{5}{*}{\textbf{Misaligned}} 
& ToM-0 vs ToM-0 &0.00 (0.00) &11.67 (20.69)&30.00 (0.00)&30.00 (0.00)&100.00 (0.00)\\
& ToM-0 vs ToM-2 &0.00 (0.00)&16.00 (26.47)&28.17 (5.64)&30.00 (0.00)&94.97 (13.59)\\
& ToM-1 vs ToM-1 &0.00 (0.00)&0.00 (0.00)&23.37 (9.38)&30.00 (0.00)&96.40 (8.89)\\
& ToM-2 vs ToM-0 &0.00 (0.00)&22.00 (27.72)&30.00 (0.00)&29.93 (0.37)&99.13 (2.94)\\
& ToM-2 vs ToM-2 &0.00 (0.00)&12.33 (23.59)&29.67 (1.37)&30.00 (0.00)&83.50 (22.03)\\
\midrule
\multirow{4}{*}{\textbf{Aligned}} 
& ToM-0 vs ToM-1 &\textbf{75.00} (0.00)&\textbf{75.00} (0.00)&6.00 (0.00)&8.13 (1.63)&44.17 (3.74)\\
& ToM-1 vs ToM-0 &\textbf{75.00} (0.00)&\textbf{75.00} (0.00)&6.03 (0.18)&\textbf{7.07} (0.37)&\textbf{43.83} (4.50)\\
& ToM-1 vs ToM-2 &\textbf{75.00} (0.00)&\textbf{75.00} (0.00)&6.10 (0.66)&7.10 (0.40)&48.90 (12.27)\\
& ToM-2 vs ToM-1 &\textbf{75.00} (0.00)&\textbf{75.00} (0.00)&\textbf{5.93} (0.37)&7.93 (1.36)&51.00 (12.88)\\
\bottomrule
\end{tabular}
\caption{
Average coordination performance between two agents with fixed ToM orders across five task settings.
Different orderings of the same agent profile (e.g., \textit{ToM-0 vs ToM-2} and \textit{ToM-2 vs ToM-0}) indicate the cases where the two agents take the roles of Player 1 and Player 2, respectively.
For repeated matrix games (Memory-1 and Memory-N), we employ the \textit{Point} obtained by the agent from the game (ranging from 0 to 75) as the metric. 
A higher point value indicates better performance.
For grid world navigation tasks (Game 1 and Game 2) and Overcooked, we employ the completion \textit{Time} (ranging from 0 to 30 for two grid world navigation task, and 0 to 100 for Overcooked) as the metric. 
For the failed samples, their completion times are setting to the maximum step limit. A lower completion time value indicates better performance. The values in parentheses indicate the variance.
}
\label{table:tom}
\end{table*}

\section{4. Experimental Setup}

\subsection{4.1. Task Setup}
We conduct empirical evaluations in three fully cooperative environments with different structures: a repeated matrix game, two grid world navigation tasks, and an Overcooked scenario.
\begin{figure}[h]
  \centering
  \begin{subfigure}{0.22\linewidth}
    \centering
    \captionsetup{justification=centering}
    \includegraphics[width=0.8\linewidth]{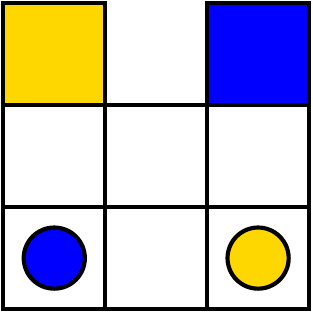}
    \caption{Game 1}
    \label{fig:sub1}
  \end{subfigure}
  \begin{subfigure}{0.36\linewidth}
    \centering
    \captionsetup{justification=centering}
    \includegraphics[width=0.8\linewidth]{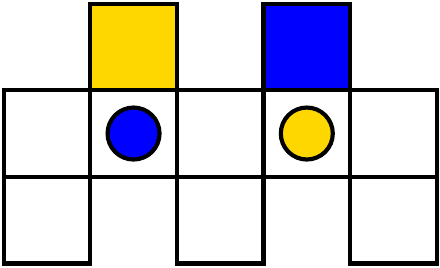}
    \caption{Game 2}
    \label{fig:sub2}
  \end{subfigure}
  \begin{subfigure}{0.33\linewidth}
     \captionsetup{justification=centering}
    \includegraphics[width=1\linewidth]{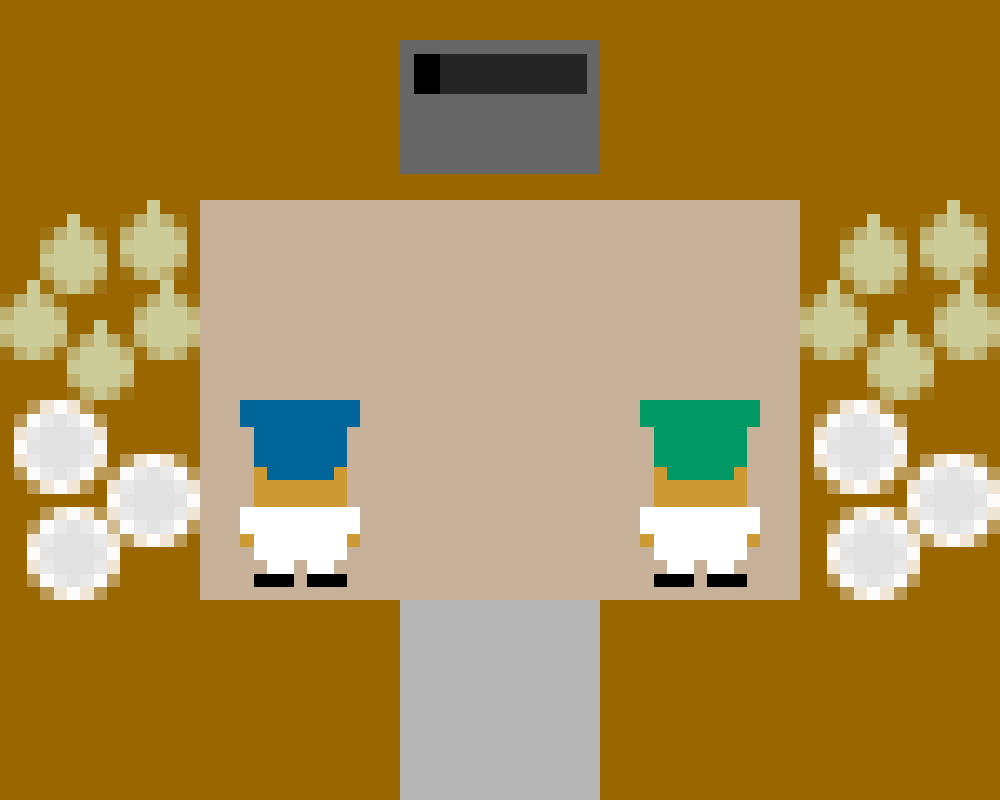}
    \caption{Overcooked}
    \label{fig:overcooked}
    \end{subfigure}
  \label{fig:envs}
  \caption{Illustrative diagrams of two grid world navigation tasks and the Overcooked layout.}
\end{figure}
\subsubsection{Repeated Matrix Game.} 
In each round, two agents select simultaneously between two options A and B without communication or prior agreement. If both agents select the same option (A-A or B-B), they each receive 0 points; if they select different options (A-B or B-A), they each receive 5 points. Therefore, agents must consistently break the symmetry in order to achieve stable coordination and maximize collective reward. We consider two settings: Memory-1 and Memory-N. In the Memory-1 setting, each agent can observe the partner’s action in the previous round. In the Memory-N setting, each agent has access to the cumulative counts of how many times the partner has chosen each option. 

\subsubsection{Grid World Navigation \citep{kleiman2016coordinate}.} Two agents are each assigned a distinct color and a corresponding target location. The goal is for both agents to reach their own target locations. At each step, both agents simultaneously move in one of four directions: up, down, left, and right. Agents cannot move outside the grid, enter the other agent's goal, occupy the same position, or swap positions within a single step. An episode ends once both agents have successfully reached their goals. We employ two grid world navigation tasks: Game 1 (\cref{fig:sub1}) and Game 2  (\cref{fig:sub2}). 
In Game 1, agents must coordinate over their trajectories to avoid blocking each other, while Game 2’s narrower layout requires one agent to temporarily move away from its own goal to make room for the partner, making coordination more challenging.


\subsubsection{Overcooked \citep{carroll2019utility}.}
In this scenario (\cref{fig:overcooked}), two agents need to collaborate to cook and deliver onion soup.
Each agent can move up, down, left, and right within the layout and interact with permissible kitchen facilities, icluding one pot, one delivery area, two onion dispensers, two plate dispensers, and four kitchen counters. 
The pot automatically cooks three onions into soup over 20 time steps. So agents need to place enough onions into the pot and be ready to plate and deliver the soup once it is cooked.
The layout we employ is adapted from the original Cramped Room in Overcooked-AI, with the key difference that all kitchen facilities in our layout are placed in a fully symmetric configuration. This symmetry imposes a greater challenge on the agents’ coordination.


\subsection{4.2. Agent Setup}
In the repeated matrix game, under the memory-1 setting, the state s is the partner’s action in the previous round; under the memory-N setting, the state is the counts of the partner’s two choices over the past N rounds.
Since agents tend to maintain coordination indefinitely once it is achieved in the repeated matrix game, we initialize each episode in an uncoordinated state to eliminate the roughly 50\% coordination success rate that would otherwise result from random initial actions. In two grid world navigation task, the state is the positions of both players. In the Overcooked task, the state includes the positions of both players, the items they are holding, as well as the statuses of the pot and kitchen counters. At each timestep, we provide the agent with the available action space from which it selects an action.

The repeated matrix game runs for 15 steps. While two grid world navigation tasks and Overcooked tasks have maximum step limits of 30 and 100, respectively. Tasks not completed within these limits are treated as failures.
All experiments use LLaMA-3.3-70B-Instruct as the underlying LLM with temperature = 0.1. 
All random seeds are set to 42.
Each configuration is repeated independently 30 times, and we report the averaged results.
\section{5. Result}

\begin{table*}[h]
\centering
\small
\begin{tabular}{llccccc}
\toprule
\multirow{2}{*}{\textbf{Algorithm}}&\multirow{2}{*}{\textbf{Agent Profile}}& \multicolumn{1}{c}{\textbf{Memory-1}} & \multicolumn{1}{c}{\textbf{Memory-N}} & \multicolumn{1}{c}{\textbf{Game 1}} & \multicolumn{1}{c}{\textbf{Game 2}}&\multicolumn{1}{c}{\textbf{Overcooked}}  \\
\cmidrule(r){3-3}
\cmidrule(r){4-4}
\cmidrule(r){5-5}
\cmidrule(r){6-6}
\cmidrule(r){7-7}
& & Point $\uparrow$ & Point $\uparrow$ & Time $\downarrow$& Time $\downarrow$ & Time $\downarrow$\\
\midrule
\multirow{5}{*}{\textbf{FTL}} 
& A-ToM vs ToM-0 &\textbf{75.00} (0.00)&\textbf{75.00} (0.00)&6.03 (0.18)&\textbf{7.00} (0.00)&45.33 (11.62)\\
& ToM-0 vs A-ToM &\textbf{75.00} (0.00)&\textbf{75.00} (0.00)&\textbf{5.87} (0.51)&7.63 (1.19)&\textbf{43.53} (4.51)\\
& A-ToM vs ToM-1 &70.00 (0.00)&70.00 (0.00)&7.80 (0.81)&10.53 (2.01)&52.17 (10.19)\\
& ToM-1 vs A-ToM &70.00 (0.00)&70.00 (0.00)&7.70 (0.88)&9.30 (0.88)&51.83 (10.04)\\
& A-ToM vs ToM-2 &\textbf{75.00} (0.00)&\textbf{75.00} (0.00)&6.03 (0.67)&7.17 (0.53)&45.30 (5.52)\\
& ToM-2 vs A-ToM &\textbf{75.00} (0.00)&\textbf{75.00} (0.00)&6.00 (0.00)&8.60 (2.18)&47.67 (6.23)\\
& A-ToM vs A-ToM &0.00 (0.00)&0.00 (0.00)&20.90 (9.03)&27.23 (5.93)&51.17 (7.36)\\
\midrule
\multirow{4}{*}{\textbf{Hedge}} 
& A-ToM vs ToM-0 &72.83 (3.39)&72.50 (3.88)&6.00 (0.00)&8.07 (0.25)&47.47 (9.13)\\
& ToM-0 vs A-ToM &73.33 (3.56)&73.00 (3.37)&6.00 (0.00)&9.40 (2.13)&46.60 (6.69)\\
& A-ToM vs ToM-1 &70.00 (4.73)&70.17 (4.82)&7.87 (0.51)&10.07 (1.55)&57.60 (10.22)\\
& ToM-1 vs A-ToM &70.00 (4.55)&70.33 (4.72)&8.23 (1.98)&10.00 (2.20)&53.17 (7.49)\\
& A-ToM vs ToM-2 &73.00 (3.11)&72.50 (4.10)&6.27 (0.74)&8.17 (0.46)&46.00 (5.79)\\
& ToM-2 vs A-ToM &72.83 (2.84)&72.33 (3.65)&6.03 (0.18)&9.47 (1.74)&47.80 (6.91)\\
& A-ToM vs A-ToM &68.17 (10.13)&64.33 (10.81)&7.60 (2.27)&8.57 (2.80)&50.53 (7.65)\\
\bottomrule
\end{tabular}
\caption{
Average coordination performance of our A-ToM agent across five task settings.
The meaning and calculation of the metrics are the same as in \Cref{table:tom}.
}
\label{table:ATA}
\end{table*}

\subsection{5.1. Effect of ToM Misalignment} 
\Cref{table:tom} presents the effects of ToM misalignment on coordination performance across different tasks and evaluation metrics. Overall, across all tasks, coordination is most successful when the agent's ToM order is aligned with that of the partner. In repeated matrix games, we observe that under the Memory-1 setting, only certain aligned pairs achieve high points. The reason lies in overthinking caused by misaligned ToM orders. For example, when two ToM-1 agents play, each assumes that the other will change their choice due to a coordination failure in the previous step. As a result, they neither updates their own choice, leading to repeated failures.
Under the Memory-N setting, the coordination pattern becomes more widespread, and misaligned pairs also exhibit some success. 
Except for ToM-1 agents in self-play, coordination failures often appear as both agents repeatedly alternating between the same two options. Over time, this back-and-forth causes both actions to occur in memory with similar frequency. Eventually, the randomness in the choice of LLM-based agent may result in occasional successful coordination.

Interestingly, we can observe that the results on the two grid world navigation tasks (especially Game 2) more closely resemble those under the Memory-1 setting in the repeated matrix game, whereas the results on Overcooked are more similar to those under the Memory-N setting.
This discrepancy may be attributed to Overcooked’s larger action space and the reduced decision optimality resulting from its higher difficulty. We will discuss this point in more detail later. Nevertheless, all this result supports our core hypothesis: in cooperative settings, alignment of ToM orders significantly enhances coordination performance, and this effect generalizes across different types of tasks.

\begin{figure}[h]
  \centering
  \includegraphics[width=0.97\linewidth]{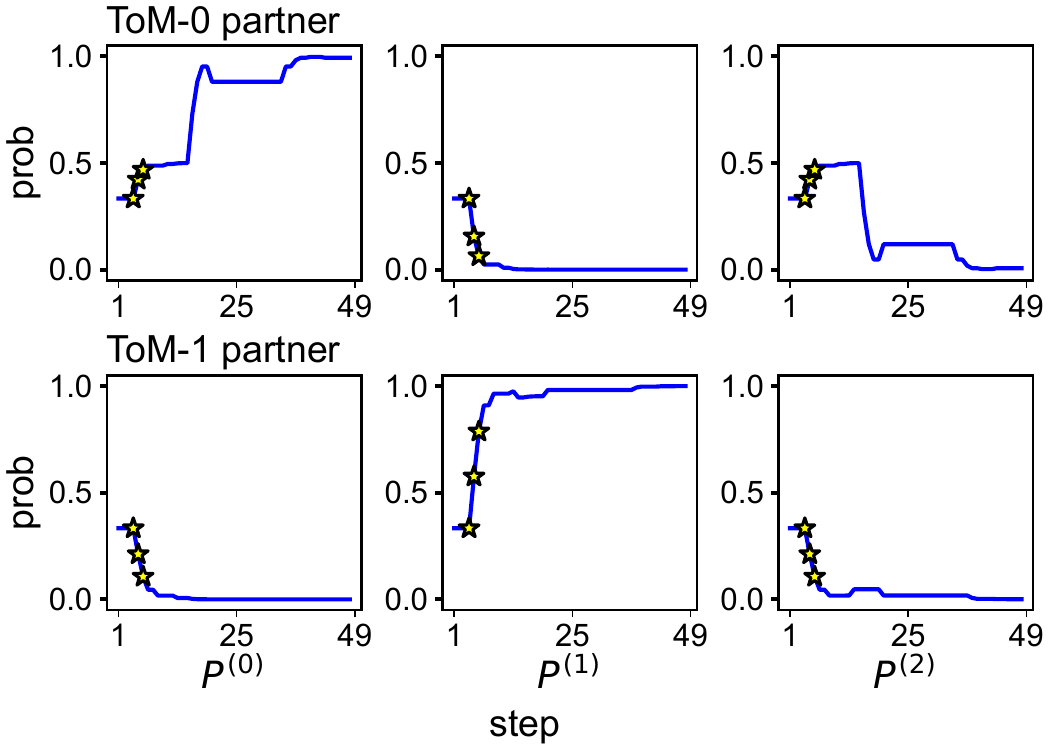}
  \caption{
    The evolution of the normalized expert weights of the A-ToM (Hedge) agents when collaborating with ToM-0 and ToM-1 partners.}
  \label{fig:tom_p}
\end{figure}

\subsection{5.2. Performance of A-ToM Agent}
\Cref{table:ATA} shows the coordination performance of adaptive agents in all tasks different interacting with fixed-ToM-order partners. Overall, for all partners with fixed ToM orders, both FTL and Hedge A-ToM agent show strong performance, as if they were the agents with ToM orders aligned to that of the partners. Nevertheless, in coordination with partners of fixed ToM orders, FTL exhibits a slight performance advantage over Hedge, likely due to its ability to more rapidly identify and adapt to the partner’s ToM characteristics.
In self-play between two A-ToM agents, however, Hedge A-ToM agent demonstrates significantly stronger adaptability than FTL, especially in repeated matrix games and two grid world navigation tasks. This is because the Hedge algorithm has a greater capacity for exploration, making it easier to align with a changing partner in terms of ToM order. In summary, the results of the two A-ToM agents are consistent with the characteristics of the respective online learning algorithms they employ, and the failure of FTL self-play highlights the importance of achieving coordination in terms of ToM order.

\begin{figure}[h]
  \centering
  \includegraphics[width=0.95\linewidth]{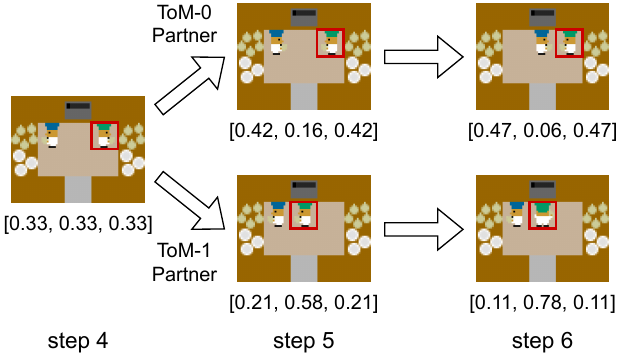}
  \caption{
  Snapshots of the game state of Overcooked at step 4, 5, and 6 when the A-ToM (Hedge) collaborating with ToM-0 and ToM-1 partners. The values below the snapshot are normalized expert weights of A-ToM agents (Hedge), which are marked with star symbols in \cref{fig:tom_p}.
}
  \label{fig:snapshot}
\end{figure}
\subsection{5.3. Case Study}
How does our A-ToM agent work in detail? We present two cases from our experiments, where the Hedge A-ToM agent plays with a ToM-0 agent and a ToM-1 agent in Overcooked. \Cref{fig:tom_p} shows how A-ToM agents's normalized expert weights ($P^{(0)}$, $P^{(1)}$, $P^{(2)}$) evolves. 
We observe that as the game progresses, the A-ToM agent gradually identifies the partner’s ToM order correctly. 
To better understand what drives the adjustment, we analyze the first update of normalized expert weights in both cases---both of which occur at step 5. 
We present snapshots of the game state at steps 4, 5, and 6 in \Cref{fig:snapshot}, and highlight these steps in \Cref{fig:tom_p}. Up to step 4, two agents' actions do not require coordination, so the A-ToM agents cannot yet infer the partner’s ToM order at that stage. However, at step 5, when both agents attempt to approach the pot, the actions differ between the ToM-0 agent and the ToM-1 agent. The ToM-0 agent does not take the A-ToM agent's action into account, resulting in a conflict with the A-ToM agent as they both try to occupy the same position. In contrast, the ToM-1 agent infers that the A-ToM agent intends to move west and thus select to stay. Based on this observation, the A-ToM agents update their expert weights at the end of this step. The updated expert weights then influence its behavior in step 6: the A-ToM agent paired with the ToM-0 agent yields to avoid conflict; the A-ToM agent paired with the ToM-1 agent prepares to place the ingredient into the pot in order to complete the task as quickly as possible.
The two cases above also help illustrate how misaligned ToM orders can lead to coordination failures.


\section{6. Generalization Analysis}
\subsection{6.1. Play with Non-LLM-Based Agents}
We now examine how the A-ToM agent collaborates with non-LLM-based partners. Using Overcooked as a case study, we consider two baseline agents: (1) Greedy, a planning-based agent implemented in the official Overcooked-ai library \citep{carroll2019utility}, and (2) PBT, a widely used standard MARL agent baseline in the Overcooked environment \citep{jaderberg2017population, carroll2019utility, 10.5555/3618408.3619252}. As shown in \cref{table:cross_play}, our A-ToM agent demonstrates the strongest generalization performance. 
Notably, the A-ToM agent tends to interpret the greedy agent and the PBT agent as ToM-0 agents in most cases, and as ToM-2 agents in a minority of cases. Specifically, the average normalized expert weights are as follows: FTL on the greedy agent---[0.93, 0.01, 0.05], FTL on the PBT agent---[0.80, 0.05, 0.16], Hedge on the greedy agent---[0.75, 0.08, 0.18], and Hedge on the PBT agent---[0.55, 0.09, 0.36]. 
Given that ToM-0 and ToM-2 agents may sometimes produce identical decisions, and considering the inherent randomness in behavior, we find that \textit{from the perspective of an A-ToM agent}, the planning or RL agent does not exhibit genuine ToM capabilities.

\begin{table}[h]
\centering
\small
\begin{tabular}{lc}
\toprule
\multirow{2}{*}{\textbf{Agent Type}}&\multicolumn{1}{c}{\textbf{Overcooked}}  \\
\cmidrule(r){2-2}
& Time $\downarrow$\\
\midrule
ToM-0 &64.60 (26.62)\\
ToM-1 &55.97 (19.80)\\
ToM-2 &60.41 (23.70)\\
Greedy &54.06 (22.47)\\
PBT &49.09 (17.63)\\
A-ToM agent (FTL) &\textbf{48.05} (10.44)\\
A-ToM agent (Hedge) &49.16 (10.83)\\
\bottomrule
\end{tabular}
\caption{
Average coordination performance of different types of agents under cross-play settings in Overcooked. The values in parentheses indicate the variance.}
\label{table:cross_play}
\end{table}

\subsection{6.2. When ToM Alignment May Not Matter}
In this paper, we highlight the importance of ToM alignment for effective coordination. However, several factors may influence this importance. As demonstrated in our earlier experiments, potential factors include the size of the task’s optimal action space and the rationality of the agents.
To validate these factors, we evaluate agents with different ToM orders in a 3-action repeated matrix game, where each agent’s action space consists of three actions: A, B, and C.
The reward rule remains the same: both players receive 5 points only when they choose different actions. What differs is that for each action taken by the partner, there are now two actions that can successfully coordinate with it. In addition, we set the LLM’s temperature to 0.9 to make its decisions less rational and its outputs more diverse.

\Cref{table:tom_importance} presents the results: compared to the results in \Cref{table:tom}, the scores between ToM-misaligned agents is higher, while the scores between ToM-aligned agents is lower. This offers preliminary evidence for our method’s applicability: \textit{the clearer the optimal action space and the more rational the agents, the greater the ToM misalignment failures—and the more effective our approach.}

\begin{table}[h]
\centering
\small
\begin{tabular}{llcc}
\toprule
\multirow{2}{*}{\textbf{Alignment}}&\multirow{2}{*}{\textbf{Agent Pair}}& \multicolumn{1}{c}{\textbf{Memory-1}} & \multicolumn{1}{c}{\textbf{Memory-N}}\\
\cmidrule(r){3-3}
\cmidrule(r){4-4}
& & Point $\uparrow$ & Point $\uparrow$ \\
\midrule
\multirow{5}{*}{\textbf{Misaligned}} 
& ToM-0 vs ToM-0 &40.50 (29.08)&35.33 (33.50)\\
& ToM-0 vs ToM-2 &64.17 (12.94)&55.33 (29.39)\\
& ToM-1 vs ToM-1 &54.33 (33.16)&50.50 (29.37)\\
& ToM-2 vs ToM-0 &66.00 (9.86)&63.33 (25.40)\\
& ToM-2 vs ToM-2 &58.83 (27.63)&50.33 (32.82)\\
\midrule
\multirow{4}{*}{\textbf{Aligned}} 
& ToM-0 vs ToM-1 &73.50 (2.98)&65.33 (20.59)\\
& ToM-1 vs ToM-0 &\textbf{74.00} (2.42)&\textbf{71.33} (6.56)\\
& ToM-1 vs ToM-2 &46.50 (32.65)&55.83 (24.36)\\
& ToM-2 vs ToM-1 &61.00 (26.31)&70.33 (5.40)\\
\bottomrule
\end{tabular}
\caption{Average coordination performance between two agents with fixed ToM orders in 3-action repeated matrix game. The values in parentheses indicate the variance. }
\label{table:tom_importance}
\end{table}

\section{7. Conclusion}
In this paper, we report our key finding: equipping agents with ToM does not necessarily improve coordination among them—only aligned ToM reasoning leads to effective collaboration. To address this, we propose an adaptive ToM agent (A-ToM agent), which frames ToM alignment as an expert advice problem. 
A-ToM agent dynamically infers the partner's ToM order and adjusts its behavior accordingly to achieve coordination. 
Conceptually, A-ToM agent transforms behavioral coordination into alignment at the order of ToM.
Through experiments across multiple tasks, we demonstrate both the validity of our finding and the effectiveness of the proposed A-ToM agent architecture. 
Furthermore, we analyze the scenarios in which ToM alignment is required the most.
We hope that our work can offer valuable insights into promoting collaboration among today's increasingly capable agents.

\section{Acknowledgments}
This research was supported in part by the National Key Research and Development Project of China (No. 2024YFE0210900), the National Science Fund for Distinguished Young Scholarship of China (No. 62025602), the National Natural Science Foundation of China (Nos. U22B2036, 6250076060 and 62506300), the Technological Innovation Team of Shaanxi Province (No. 2025RS-CXTD-009), the International Cooperation Project of Shaanxi Province (No. 2025GH-YBXM-017), the Shanghai Municipal Science and Technology Major Project, the Tencent Foundation and Xplorer Prize.
\bibliography{aaai2026}

@inproceedings{qian2024chatdev,
  title={ChatDev: Communicative Agents for Software Development},
  author={Qian, Chen and Liu, Wei and Liu, Hongzhang and Chen, Nuo and Dang, Yufan and Li, Jiahao and Yang, Cheng and Chen, Weize and Su, Yusheng and Cong, Xin and others},
  booktitle={Proceedings of the 62nd Annual Meeting of the Association for Computational Linguistics (Volume 1: Long Papers)},
  pages={15174--15186},
  year={2024}
}

@inproceedings{hong2023metagpt,
  title={MetaGPT: Meta programming for a multi-agent collaborative framework},
  author={Hong, Sirui and Zhuge, Mingchen and Chen, Jonathan and Zheng, Xiawu and Cheng, Yuheng and Wang, Jinlin and Zhang, Ceyao and Wang, Zili and Yau, Steven Ka Shing and Lin, Zijuan and others},
  booktitle={The Twelfth International Conference on Learning Representations},
  year={2023}
}

@article{agashe2023llm,
  title={Llm-coordination: evaluating and analyzing multi-agent coordination abilities in large language models},
  author={Agashe, Saaket and Fan, Yue and Reyna, Anthony and Wang, Xin Eric},
  journal={arXiv preprint arXiv:2310.03903},
  year={2023}
}

@article{cross2024hypothetical,
  title={Hypothetical minds: Scaffolding theory of mind for multi-agent tasks with large language models},
  author={Cross, Logan and Xiang, Violet and Bhatia, Agam and Yamins, Daniel LK and Haber, Nick},
  journal={arXiv preprint arXiv:2407.07086},
  year={2024}
}

@article{wellman2018theory,
  title={Theory of mind: The state of the art},
  author={Wellman, Henry M},
  journal={European Journal of Developmental Psychology},
  volume={15},
  number={6},
  pages={728--755},
  year={2018},
  publisher={Taylor \& Francis}
}

@inproceedings{de2014agent,
  title={Agent-based models for higher-order theory of mind},
  author={de Weerd, Harmen and Verbrugge, Rineke and Verheij, Bart},
  booktitle={Advances in Social Simulation: Proceedings of the 9th Conference of the European Social Simulation Association},
  pages={213--224},
  year={2014},
  organization={Springer}
}

@article{shao2024cognitive,
  title={Cognitive Insights and Stable Coalition Matching for Fostering Multi-Agent Cooperation},
  author={Shao, Jiaqi and Yuan, Tianjun and Lin, Tao and Cao, Xuanyu and Luo, Bing},
  journal={arXiv preprint arXiv:2405.18044},
  year={2024}
}

@article{cesa1997use,
  title={How to use expert advice},
  author={Cesa-Bianchi, Nicolo and Freund, Yoav and Haussler, David and Helmbold, David P and Schapire, Robert E and Warmuth, Manfred K},
  journal={Journal of the ACM (JACM)},
  volume={44},
  number={3},
  pages={427--485},
  year={1997},
  publisher={ACM New York, NY, USA}
}

@article{yoshida2008game,
  title={Game theory of mind},
  author={Yoshida, Wako and Dolan, Ray J and Friston, Karl J},
  journal={PLoS computational biology},
  volume={4},
  number={12},
  pages={e1000254},
  year={2008},
  publisher={Public Library of Science San Francisco, USA}
}

@article{de2013much,
  title={How much does it help to know what she knows you know? An agent-based simulation study},
  author={De Weerd, Harmen and Verbrugge, Rineke and Verheij, Bart},
  journal={Artificial Intelligence},
  volume={199},
  pages={67--92},
  year={2013},
  publisher={Elsevier}
}

@inproceedings{wangtom2c,
  title={ToM2C: Target-oriented Multi-agent Communication and Cooperation with Theory of Mind},
  author={Wang, Yuanfei and Xu, Jing and Wang, Yizhou and others},
  booktitle={International Conference on Learning Representations},
  year={2022}
}

@article{fuchs2021theory,
  title={Theory of mind for deep reinforcement learning in hanabi},
  author={Fuchs, Andrew and Walton, Michael and Chadwick, Theresa and Lange, Doug},
  journal={arXiv preprint arXiv:2101.09328},
  year={2021}
}

@article{wu2021too,
  title={Too many cooks: Bayesian inference for coordinating multi-agent collaboration},
  author={Wu, Sarah A and Wang, Rose E and Evans, James A and Tenenbaum, Joshua B and Parkes, David C and Kleiman-Weiner, Max},
  journal={Topics in Cognitive Science},
  volume={13},
  number={2},
  pages={414--432},
  year={2021},
  publisher={Wiley Online Library}
}

@inproceedings{li2023theory,
  title={Theory of Mind for Multi-Agent Collaboration via Large Language Models},
  author={Li, Huao and Chong, Yu and Stepputtis, Simon and Campbell, Joseph P and Hughes, Dana and Lewis, Charles and Sycara, Katia},
  booktitle={Proceedings of the 2023 Conference on Empirical Methods in Natural Language Processing},
  pages={180--192},
  year={2023}
}

@inproceedings{zhang2024proagent,
  title={ProAgent: building proactive cooperative agents with large language models},
  author={Zhang, Ceyao and Yang, Kaijie and Hu, Siyi and Wang, Zihao and Li, Guanghe and Sun, Yihang and Zhang, Cheng and Zhang, Zhaowei and Liu, Anji and Zhu, Song-Chun and others},
  booktitle={Proceedings of the AAAI Conference on Artificial Intelligence},
  volume={38},
  number={16},
  pages={17591--17599},
  year={2024}
}

@inproceedings{brohan2023can,
  title={Do as i can, not as i say: Grounding language in robotic affordances},
  author={Brohan, Anthony and Chebotar, Yevgen and Finn, Chelsea and Hausman, Karol and Herzog, Alexander and Ho, Daniel and Ibarz, Julian and Irpan, Alex and Jang, Eric and Julian, Ryan and others},
  booktitle={Conference on robot learning},
  pages={287--318},
  year={2023},
  organization={PMLR}
}

@article{wu2023tidybot,
  title={Tidybot: Personalized robot assistance with large language models},
  author={Wu, Jimmy and Antonova, Rika and Kan, Adam and Lepert, Marion and Zeng, Andy and Song, Shuran and Bohg, Jeannette and Rusinkiewicz, Szymon and Funkhouser, Thomas},
  journal={Autonomous Robots},
  volume={47},
  number={8},
  pages={1087--1102},
  year={2023},
  publisher={Springer}
}

@inproceedings{xia2023towards,
  title={Towards autonomous system: flexible modular production system enhanced with large language model agents},
  author={Xia, Yuchen and Shenoy, Manthan and Jazdi, Nasser and Weyrich, Michael},
  booktitle={2023 IEEE 28th International Conference on Emerging Technologies and Factory Automation (ETFA)},
  pages={1--8},
  year={2023},
  organization={IEEE}
}

@inproceedings{gurreal,
  title={A Real-World WebAgent with Planning, Long Context Understanding, and Program Synthesis},
  author={Gur, Izzeddin and Furuta, Hiroki and Huang, Austin V and Safdari, Mustafa and Matsuo, Yutaka and Eck, Douglas and Faust, Aleksandra},
  booktitle={The Twelfth International Conference on Learning Representations},
  year={2024}
}

@article{yan2023gpt,
  title={Gpt-4v in wonderland: Large multimodal models for zero-shot smartphone gui navigation},
  author={Yan, An and Yang, Zhengyuan and Zhu, Wanrong and Lin, Kevin and Li, Linjie and Wang, Jianfeng and Yang, Jianwei and Zhong, Yiwu and McAuley, Julian and Gao, Jianfeng and others},
  journal={arXiv preprint arXiv:2311.07562},
  year={2023}
}

@article{wang2023voyager,
  title={Voyager: An open-ended embodied agent with large language models},
  author={Wang, Guanzhi and Xie, Yuqi and Jiang, Yunfan and Mandlekar, Ajay and Xiao, Chaowei and Zhu, Yuke and Fan, Linxi and Anandkumar, Anima},
  journal={arXiv preprint arXiv:2305.16291},
  year={2023}
}

@inproceedings{wang2023describe,
  title={Describe, explain, plan and select: interactive planning with large language models enables open-world multi-task agents},
  author={Wang, Zihao and Cai, Shaofei and Chen, Guanzhou and Liu, Anji and Ma, Xiaojian and Liang, Yitao and CraftJarvis, Team},
  booktitle={Proceedings of the 37th International Conference on Neural Information Processing Systems},
  pages={34153--34189},
  year={2023}
}

@article{camerer2004cognitive,
  title={A cognitive hierarchy model of games},
  author={Camerer, Colin F and Ho, Teck-Hua and Chong, Juin-Kuan},
  journal={The Quarterly Journal of Economics},
  volume={119},
  number={3},
  pages={861--898},
  year={2004},
  publisher={MIT Press}
}

@article{devaine2014social,
  title={The social Bayesian brain: does mentalizing make a difference when we learn?},
  author={Devaine, Marie and Hollard, Guillaume and Daunizeau, Jean},
  journal={PLoS computational biology},
  volume={10},
  number={12},
  pages={e1003992},
  year={2014},
  publisher={Public Library of Science San Francisco, USA}
}

@article{kalai2005efficient,
  title={Efficient algorithms for online decision problems},
  author={Kalai, Adam and Vempala, Santosh},
  journal={Journal of Computer and System Sciences},
  volume={71},
  number={3},
  pages={291--307},
  year={2005},
  publisher={Elsevier}
}

@article{freund1997decision,
  title={A decision-theoretic generalization of on-line learning and an application to boosting},
  author={Freund, Yoav and Schapire, Robert E},
  journal={Journal of computer and system sciences},
  volume={55},
  number={1},
  pages={119--139},
  year={1997},
  publisher={Elsevier}
}

@inproceedings{kleiman2016coordinate,
  title={Coordinate to cooperate or compete: abstract goals and joint intentions in social interaction},
  author={Kleiman-Weiner, Max and Ho, Mark K and Austerweil, Joseph L and Littman, Michael L and Tenenbaum, Joshua B},
  booktitle={Proceedings of the annual meeting of the cognitive science society},
  volume={38},
  year={2016}
}

@inproceedings{boutilier1999sequential,
  title={Sequential optimality and coordination in multiagent systems},
  author={Boutilier, Craig},
  booktitle={IJCAI},
  volume={99},
  pages={478--485},
  year={1999}
}

@article{carroll2019utility,
  title={On the utility of learning about humans for human-ai coordination},
  author={Carroll, Micah and Shah, Rohin and Ho, Mark K and Griffiths, Tom and Seshia, Sanjit and Abbeel, Pieter and Dragan, Anca},
  journal={Advances in neural information processing systems},
  volume={32},
  year={2019}
}

@inproceedings{park2023generative,
  title={Generative agents: Interactive simulacra of human behavior},
  author={Park, Joon Sung and O'Brien, Joseph and Cai, Carrie Jun and Morris, Meredith Ringel and Liang, Percy and Bernstein, Michael S},
  booktitle={Proceedings of the 36th annual acm symposium on user interface software and technology},
  pages={1--22},
  year={2023}
}

@inproceedings{zhu2021few,
  title={Few-shot language coordination by modeling theory of mind},
  author={Zhu, Hao and Neubig, Graham and Bisk, Yonatan},
  booktitle={International conference on machine learning},
  pages={12901--12911},
  year={2021},
  organization={PMLR}
}

@inproceedings{10.5555/3618408.3619252,
author = {Li, Yang and Zhang, Shao and Sun, Jichen and Du, Yali and Wen, Ying and Wang, Xinbing and Pan, Wei},
title = {Cooperative Open-Ended Learning Framework for Zero-Shot Coordination},
year = {2023},
publisher = {JMLR.org},
booktitle = {Proceedings of the 40th International Conference on Machine Learning},
articleno = {844},
numpages = {15},
location = {Honolulu, Hawaii, USA},
series = {ICML'23}
}

@article{jaderberg2017population,
  title={Population based training of neural networks},
  author={Jaderberg, Max and Dalibard, Valentin and Osindero, Simon and Czarnecki, Wojciech M and Donahue, Jeff and Razavi, Ali and Vinyals, Oriol and Green, Tim and Dunning, Iain and Simonyan, Karen and others},
  journal={arXiv preprint arXiv:1711.09846},
  year={2017}
}

@inproceedings{zhang2025k,
  title={K-Level Reasoning: Establishing Higher Order Beliefs in Large Language Models for Strategic Reasoning},
  author={Zhang, Yadong and Mao, Shaoguang and Ge, Tao and Wang, Xun and Xia, Yan and Lan, Man and Wei, Furu},
  booktitle={Proceedings of the 2025 Conference of the Nations of the Americas Chapter of the Association for Computational Linguistics: Human Language Technologies (Volume 1: Long Papers)},
  pages={7212--7234},
  year={2025}
}

@article{zhang2024multi,
  title={Multi-agent reinforcement learning for autonomous driving: A survey},
  author={Zhang, Ruiqi and Hou, Jing and Walter, Florian and Gu, Shangding and Guan, Jiayi and R{\"o}hrbein, Florian and Du, Yali and Cai, Panpan and Chen, Guang and Knoll, Alois},
  journal={arXiv preprint arXiv:2408.09675},
  year={2024}
}

@article{kegeleirs2025towards,
  title={Towards applied swarm robotics: current limitations and enablers},
  author={Kegeleirs, Miquel and Birattari, Mauro},
  journal={Frontiers in Robotics and AI},
  volume={12},
  pages={1607978},
  year={2025},
  publisher={Frontiers Media SA}
}

@article{ge2025distributed,
  title={Distributed coordination control of multi-agent systems under intermittent sampling and communication: a comprehensive survey},
  author={Ge, Xiaohua and Han, Qing-Long and Zhang, Xian-Ming and Ding, Derui and Ning, Boda},
  journal={Science China Information Sciences},
  volume={68},
  number={5},
  pages={151201},
  year={2025},
  publisher={Springer}
}

@inproceedings{liu2024llm,
  title={LLM-Powered Hierarchical Language Agent for Real-time Human-AI Coordination},
  author={Liu, Jijia and Yu, Chao and Gao, Jiaxuan and Xie, Yuqing and Liao, Qingmin and Wu, Yi and Wang, Yu},
  booktitle={Proceedings of the 23rd International Conference on Autonomous Agents and Multiagent Systems},
  pages={1219--1228},
  year={2024}
}

@inproceedings{hu2020other,
  title={“other-play” for zero-shot coordination},
  author={Hu, Hengyuan and Lerer, Adam and Peysakhovich, Alex and Foerster, Jakob},
  booktitle={International Conference on Machine Learning},
  pages={4399--4410},
  year={2020},
  organization={PMLR}
}

\end{document}